\documentclass[twocolumn]{article}
\usepackage{graphicx, anysize, xurl, color}
\usepackage{makecell,hyperref}
\usepackage{csquotes,xspace}
\marginsize{0.70in}{0.70in}{0.70in}{0.75in}

\newcommand{\eat}[1]{}

\newcommand{\blue}[1]{{\color{blue} #1}}

\title{
The False Dawn: Reevaluating Google's Reinforcement Learning \\ for Chip Macro Placement
}
\author{Igor L. Markov, igor.markov@gmail.com}
\date{}

\begin{document}
\maketitle

\begin{abstract}
Reinforcement learning (RL) for physical design of silicon chips in a Google 2021 Nature paper stirred controversy due to poorly documented claims that raised eyebrows and  drew critical media coverage. The paper withheld critical methodology steps along with most inputs needed to reproduce results. Our meta-analysis shows how two separate evaluations filled in the gaps and demonstrated that Google RL lags behind ($i$) human chip designers, ($ii$) a well-known algorithm (Simulated Annealing), and ($iii$) generally-available commercial software, while being slower; and in a 2023 open research contest, RL methods weren't in top 5. Crosschecked data indicate that the integrity of the Nature paper is substantially undermined owing to errors in conduct, analysis and reporting. Before publishing, Google rebuffed internal allegations of fraud. We note policy implications and conclusions for chip design.
\end{abstract}

\noindent {\bf About the Author:}
Dr. Igor L. Markov worked on Physical Design of Integrated Circuits for over 25 years~\cite{Capo,Benchmarks,SimPL,Progress}, coauthored dozens of highly-cited peer-reviewed papers in the field, as well as a Physical Design textbook \cite{PDtextbook}. During his tenure at the Univ. of Michigan, he supervised 3 Ph.D. dissertations on chip floorplanning and circuit placement. Dr. Markov coauthored algorithms and software packages for chip floorplanning and circuit placement that were used to design many commercial chips.
He served as an editor of top EDA journals and of the two-volume Handbook of EDA \cite{EDA}. He is an IEEE Fellow and an ACM Distinguished Scientist, a Nature author and reviewer. Dr. Markov previously worked on Search at Google and ML platforms at Meta, and taught chip design at Stanford. He is a Distinguished Architect in AI \& Innovation at Synopsys and a director at
the humanitarian relief nonprofit Nova Ukraine.

This work started when Dr.Markov was affiliated with the Univ. of Michigan, but opinions given here are the author's and do not represent his employers.

{\small For a successful technology, reality must take precedence over public relations, for Nature cannot be fooled}.
--- {\small Richard Feynman, the Challenger Accident Report}

\section{Introduction}
As AI applications demand greater compute power, efficiency may be improved
via better chip design. The Nature paper \cite{Nature} by Google researchers, published in June 2021, was advertised as a chip-design breakthrough usin Machine Learning (ML) \cite{NatureEditorial2021}. It addressed a challenging problem to optimize locations of circuit components on a chip and described applications to five Tensor Processing Unit (TPU) chip blocks, implying that no better methods were available at the time in academia or industry. The paper broadened the claims beyond chip design to suggest that Reinforcement Learning (RL) extends state of the art in combinatorial optimization~\cite{RLforCO2020}.

 ``Extraordinary claims require extraordinary evidence''~\cite{False,SaganWiki}, but \cite{Nature} showed no results on public test examples (benchmarks~\cite{Benchmarks}) and did not share the proprietary TPU chip blocks used. Source code, released 7 months after publication \cite{CT} to support \cite{Nature}
after initial controversy~\cite{NYT,Reuters,Wired,FAC,TheRegister},
was missing key parts needed to reproduce the methods and results (as explained in \cite{ISPD2023,CACM}). Over a dozen researchers \cite{NYT, Reuters, TheRegister, CACM} from Google and academia questioned the claims of \cite{Nature}, performed experiments and raised concerns~\cite{SB,ISPD2023} about \cite{Nature}.

Confusingly, the then-head of Google Brain, Dr. Zoubin Ghahramani, a Google VP, \href{https://twitter.com/ZoubinGhahrama1/status/1512203509646741507?t=9kyvWT0gCXQ5coWCG7LvsA}{\blue{tweeted}}
on April 7, 2022 \cite{ZoubinTweet}
``Google stands by this work published in Nature on ML for Chip Design, which has been independently replicated, open-sourced, and used in production at Google,''
apparently referring to reproduction by another Google team (Sergio Guadarrama's), and without specifying what aspects were reproduced. Google engineers updated their open source \cite{CT} many times since, filling in some missing pieces but not all \cite{ISPD2023}.  A single open-source chip-design example was added to \cite{CT}, but results on it were neither sufficient nor clearly supportive of Google's RL code  \cite{ISPD2023}. Apparently, the only openly claimed independent (of Google) reproduction of techniques in \cite{Nature} was developed in Fall 2022 by UCSD researchers~\cite{ISPD2023}.\footnote{
Efforts by Prof. Andrew B. Kahng at UCSD were praised by Dr. Jeff Dean (the most senior author of the Nature paper~\cite{Nature} and then a Google SVP) in his recorded Dec 2, 2022 workshop keynote~\cite{JD}. When UCSD efforts were starting, Prof. Kahng publicly stated \cite{IEEESpectrum} that he was Reviewer \#3 of \cite{Nature}. In the 1990s, Prof. Kahng supervised the doctoral dissertation of the author of this meta-analysis on large-scale VLSI placement at UCLA.}
They reverse-engineered key components missing from \cite{CT} and completely reimplemented the Simulated Annealing (SA) baseline \cite{ISPD2023} absent in \cite{CT}.
Google released no proprietary TPU chip design blocks used in \cite{Nature} (nor sanitized equivalents), ruling out full external reproduction of results. So, the UCSD team shared
\cite{MP} their experiments on modern public chip designs: SA and commercial EDA tools outperformed the Google RL code \cite{CT}.

Reporters from the New York Times and Reuters covered this controversy in 2022 \cite{NYT, Reuters} and found that, well before the Nature submission, several Google researchers disputed the claims they had been tasked with checking. Two lead authors of \cite{Nature} complained of persistent allegations of fraud in their research \cite{Wired}. In 2022 Google fired an internal whistleblower \cite{NYT, Reuters} and denied publication approval for a paper~\cite{SB} written by Google researchers critical of \cite{Nature}. The whistleblower
sued Google for wrongful termination under California whistleblower-protection laws: court documents \cite{FAC}, filed under penalty of perjury, detail allegations of fraud and scientific misconduct related to research in \cite{Nature}. The lawsuit moved ahead in Aug `23 \cite{BloombergJuly23,TheRegJuly23,RulingAug23}.
Within months of the 2022 media investigations and the lawsuit, the two lead authors of \cite{Nature} and a senior coauthor left Google \cite{RichardHo}. Spring 2023 media coverage noted alleged misrepresentations by Google to potential cloud-services customers \cite{TheRegister}, questioned reproducibility of results in the Nature paper \cite{CACM}, and covered UCSD research trying to settle the dispute \cite{IEEESpectrum}. The 2021 Nature News \& Views article introducing \cite{Nature} in the same issue urged replication of results of \cite{Nature}. Given the obstacles to and the results of replication attempts \cite{ISPD2023}, the author of the article retracted it.

In this work, Section \ref{sec:background} reviews background and the chip-design task solved in \cite{Nature,CT}, then introduces secondary sources used \cite{ISPD2022,SB,MP,ISPD2023}. Section \ref{sec:initial} lists initial suspicions about \cite{Nature}. Section \ref{sec:additional} shows that many of them were confirmed later. Section \ref{sec:SOTA} checks if \cite{Nature} improved the State of the Art (SOTA). Section \ref{sec:responses} outlines how authors of \cite{Nature} responded to critiques. Section \ref{sec:use} discusses possible uses of the work in \cite{Nature} in practice. Section \ref{sec:conclusions} draws conclusions and notes policy implications.

\begin{table*}[t]
\begin{center}
\begin{tabular}{l|l|l}
Google Team 1 & Google Team 2 & UCSD Team \\
(Nature authors + coauthors)
& + external coauthors  &  \\
\hline
\hline
{\small Circuit Training (CT) repo \& FAQ} \cite{CT} & Stronger
Baselines \cite{SB} & {\small MacroPlacement repo \& FAQ} \cite{MP} \\
ISPD 2022 paper \cite{ISPD2022} & &
ISPD 2023 paper \cite{ISPD2023} \\
\hline
4 proprietary TPU blocks &  20 proprietary TPU blocks
&  All with numerous macros: \\
~~~~~~ (\cite[Figure 3]{Nature})
       &  17 public IBM circuits  \cite{ICCAD04} & 17 public IBM circuits  \cite{ICCAD04} \\
ariane (public) \cite{CT} --- &  &  $2\times$ ariane (public) \cite{MP,ISPD2023}\\
   all with numerous macros   &  all with numerous macros &  $2\times$ MemPool (public) \cite{MP,ISPD2023} \\
       &   &  $2\times$ BlackParrot (public) \cite{MP,ISPD2023} \\
\hline
\end{tabular}
\vspace{-2mm}
\caption{\label{tab:refs}
Secondary sources published by the teams and chip designs for which they report results. The IBM circuits \cite{ICCAD04} are ICCAD 2004 benchmarks. \cite{ISPD2023} built 3 designs with 2 semiconductor technologies each.
}
\end{center}
\vspace{-5mm}
\end{table*}

\section{Background}
\label{sec:background}
\vspace{-2mm}

\begin{figure*}[t]
   \vspace{-3mm}
    \centering
    \includegraphics[width=0.45\textwidth]
    {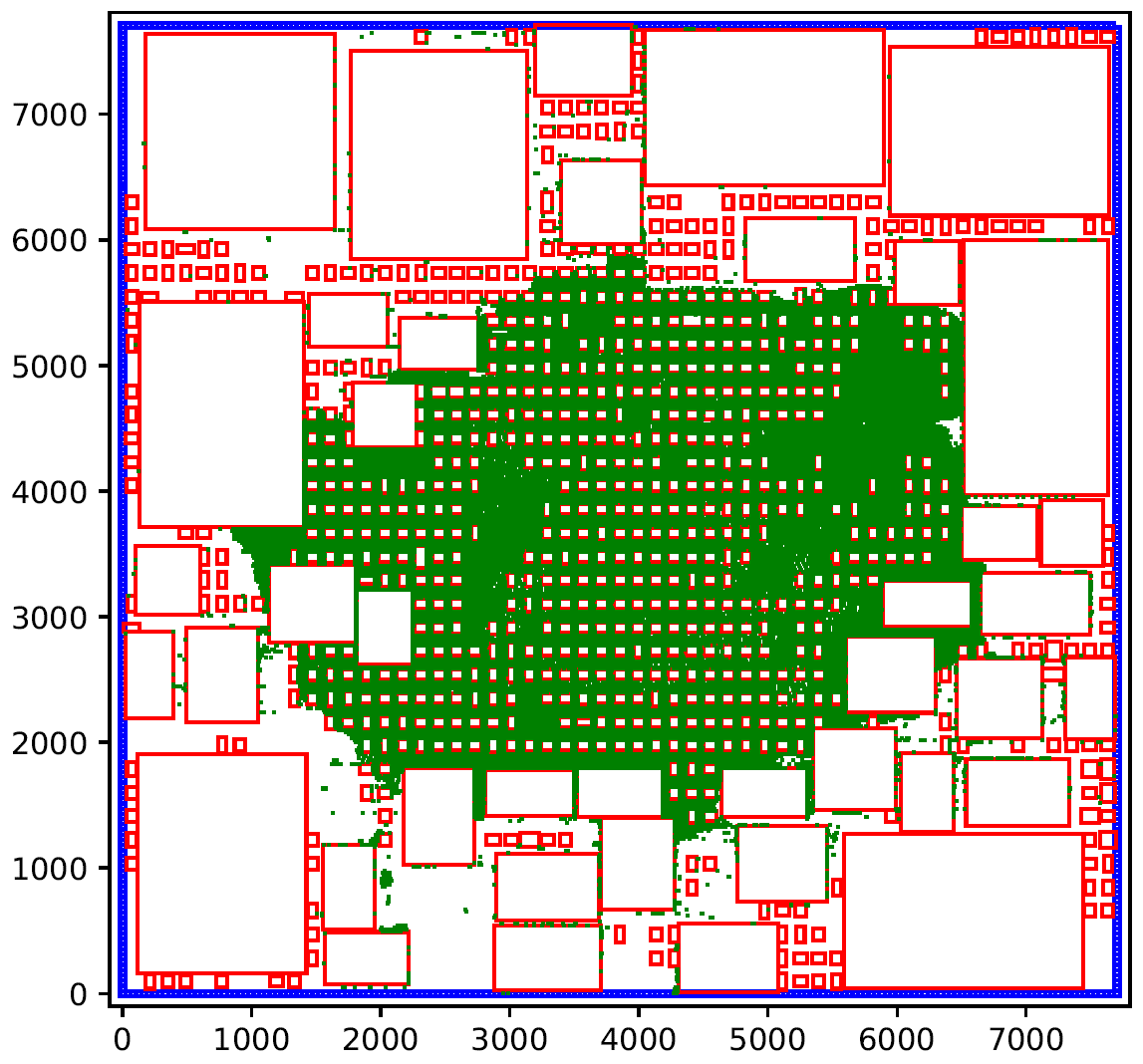}
    \includegraphics[width=0.45\textwidth]    {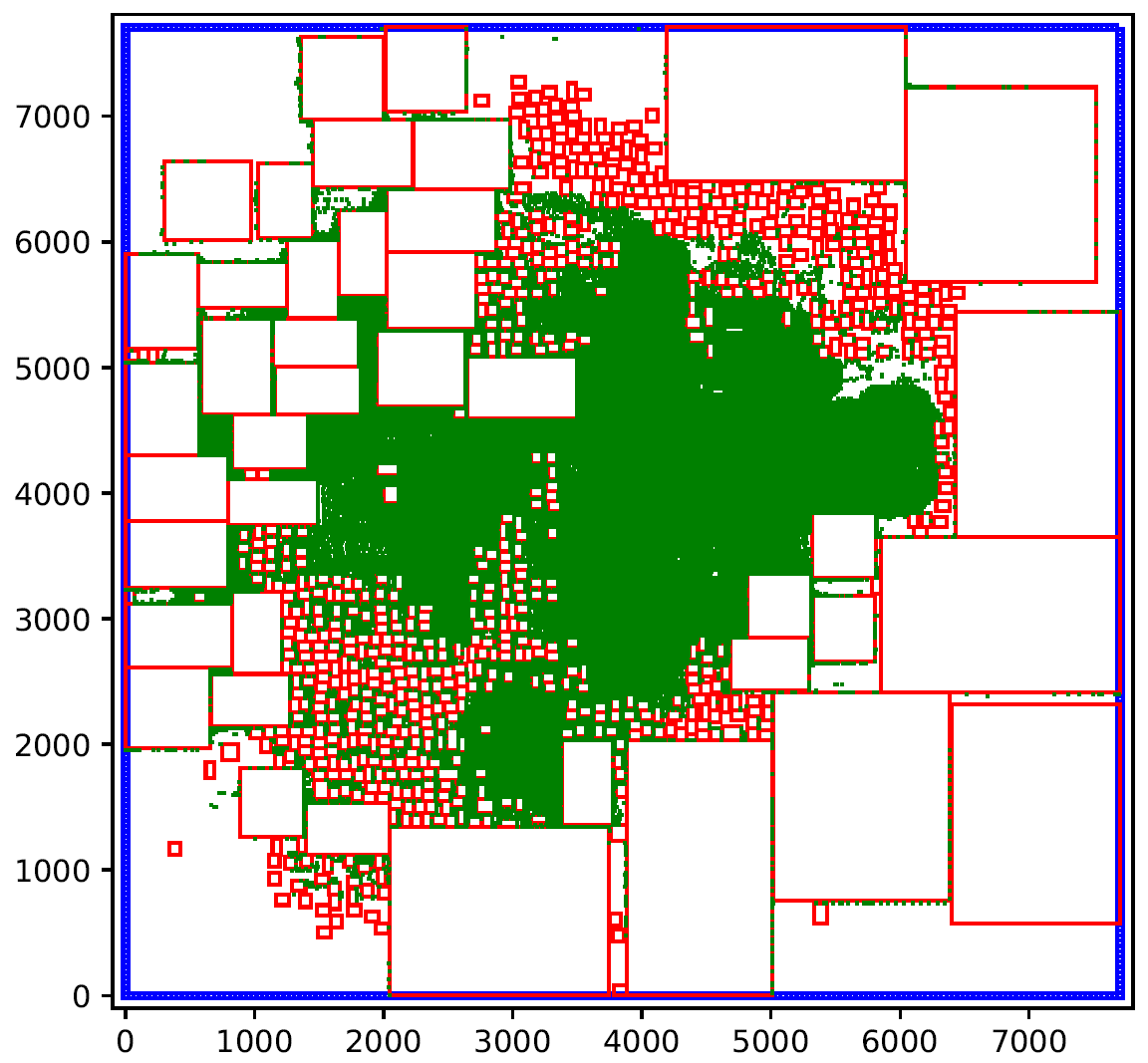}
   \vspace{-4mm}
    \caption{
     \label{fig:ibm10}
     Layouts from \cite[Figure 2]{SB} with macros in red and standard cells in green, locations produced by RL (left) and RePlAce (right) for the {\bf ibm10} benchmark from \cite{ICCAD04}. Limiting macro locations to a coarse grid (left) leads to spreading of small macros (red squares on the grid) and elongates connecting wires: from 27.5 units (right) to 44.1 units (left) for {\bf ibm10} \cite[Table 1]{SB}).
     Higher area utilization and many macros of different sizes
     make the ICCAD 2004 benchmarks \cite{ICCAD04} challenging compared to benchmarks in \cite{Nature} and \cite[page 43]{PeerReviews}.}
     \vspace{-2mm}
\end{figure*}

Components of integrated circuits include small {\em gates} and {\em standard cells}, as well as {\em memory arrays} and reusable subcircuits.
In physical design   \cite{EDA,PDtextbook}, they are modeled by rectangles within the chip canvas (Figure \ref{fig:ibm10}).
Connections between components are modeled by the {\em circuit netlist} before wire routes are known: a netlist is an unordered set of {\em nets}, each naming components that should be connected. The length of a net depends on components' locations and on wire routes; long routes are undesirable. The {\em macro placement} problem addressed in \cite{Nature} seeks $(x,y)$ locations for large circuit components ({\em macros}) so that their rectangles do not overlap and the remaining components can be placed well to optimize the chip layout \cite{CongNam,Progress,ABKplacement}.

\noindent
{\bf Circuit placement as an optimization task}.
After $(x,y)$ locations of all components are known, wires that connect components' I/O pins are routed. Routes impact chip metrics (for power, timing/speed, etc). The optimization of $(x,y)$ locations starts with simplified estimates of wirelength without wire routes. Pin locations $(x_1,y_1)$ and $(x_2,y_2)$ may be connected by horizontal and vertical wire segments in many ways, but the shortest route length is $|x_1 - x_2| + |y_1 - y_2|$. For multiple pin locations $\{(x_i,y_i)\}_i$, this estimate
generalizes to
\vspace{-2mm}
\begin{equation}
\label{eq:HPWL}
\vspace{-2mm}
\textrm{HPWL} = (\max_i  x_i - \min_i x_i) +
(\max_i y_i - \min_i y_i)
\vspace{-2mm}
\end{equation}
HPWL stands for {\em half-perimeter wirelength}, where the {\em perimeter} is taken of the {\em bounding box} of points $\{(x_i,y_i)\}_i$~\cite{CongNam,Progress,PDtextbook}. It is easy to compute and sum over many nets. This sum correlates with total routed wirelength reasonably well. \eat{HPWL is invariant under shifts of $(x,y)$ locations.} When $(x,y)$ locations are scaled by a factor $\gamma>0$, HPWL also scales by $\gamma$. This makes HPWL optimization scale-invariant and appropriate for all semiconductor technology nodes.\footnote{With semiconductor technology scaling, some macros may scale differently from logic circuits, but placement algorithms should handle a variety of macro sizes.} Algorithms that optimize HPWL extend to more precisely optimize routed wirelength and technology-dependent chip metrics, so HPWL optimization is a precursor~
\cite{Capo,Routable04,PDhandbook,NTU2012,Cong2013,ICCAD14contest,ICCAD15contestA,ICCAD15contestB,Progress,ICCAD16contest,NTU2017,RePlAce,ABKplacement}:%
\begin{itemize}
\vspace{-1mm}
\item to test new placement methods; once HPWL results are close to the best known, more accurate metrics are used for evaluation; or%
\vspace{-2mm}
\item followed by optimizations of sophisticated objectives that include HPWL (such as the proxy cost function used by RL in \cite{Nature}).
\vspace{-2mm}
\end{itemize}

 Widely adopted optimizations for placement do not use ML \cite{CongNam,PDhandbook,Progress,PDtextbook,ABKplacement} and can be classified as: ($i$) Simulated Annealing (SA), ($ii$) partitioning-driven, and ($iii$) analytical. SA, developed in the 1980s~\cite{SA83,TW85,ParAnneal87} and dominant through the mid-1990s~\cite{SAbook88,TW95,VPR}, starts with an initial layout (e.g., random) and alters it by a sequence of actions, such as component moves and swaps, of prescribed length.
 To improve the end result, some actions may sacrifice quality to escape local minima. SA excels on smaller layouts (up to 100K components) but takes a long time for large layouts. Partitioning-driven methods~\cite{Capo,FSmixed,FS5,Floorplacement,Combinatorial}
 view the circuit connectivity (the netlist) as a hypergraph and use established software packages to subdivide it into partitions with more connections within the partitions (not between). These methods run faster than SA, capture global netlist structures, and were dominant for some 10 years. Yet, the mismatch between partitioning and placement objectives (Equation \ref{eq:HPWL}) leaves room for improvement~\cite{Combinatorial}. Analytical methods approximate Equation \ref{eq:HPWL} by closed-form functions amenable to established optimization methods. {\em Force-directed placement} \cite{FD84} from the 1980s models nets by springs and finds component locations to balance out spring forces \cite{PDtextbook}. In the 2000s, advanced analytical placement techniques attained superiority \cite{CongNam,Progress,RePlAce,ABKplacement} on all large modern public benchmark sets, including those with macros and routing data~
 \cite{PEKO-MS, CRISP,MAPLE,NTU2012,Cong2013,RePlAce}. RePlAce \cite{RePlAce} from UCSD is much faster than SA and partitioning methods, but lags in quality on small netlists.

The Nature paper \cite{Nature} focuses on large circuit components ({\em macros}) among numerous small components. The {\em fixed-outline macro-placement} problem was formulated in the early 2000s \cite{FPharmful,Parquet,HardMacros}; it places all components onto a fixed-size canvas (prior formulations could stretch the canvas). it is now viewed as part of {\em mixed-size placement}
\cite{FSmixed,Combinatorial} where all  components are placed on a fixed-size canvas. A 2004 benchmark suite \cite{ICCAD04} for testing mixed-size placement algorithms evaluates the HPWL objective (Equation \ref{eq:HPWL}) which, as noted above, is apt for all semiconductor technology nodes. The suite has enjoyed significant use in the literature, e.g., \cite{Routable04,FSmixed,FS5,Combinatorial,RePlAce}.

Commercial and academic software for placement is developed to run on modest hardware within reasonable runtime. The methods and software in \cite{Nature} consume much greater resources, but at least with Simulated Annealing (during comparisons) it is straightforward to obtain progressively better results with greater runtime budget.

Circuit metrics for evaluating optimization results include circuit timing and dynamic power. Unlike power, timing metrics are sensitive to long/slow paths taken by signal transitions in a circuit and are difficult to predict before detailed placement and wire routing. Accurate early estimation of circuit metrics is a popular topic in the research literature, but remains an unsolved challenge in physical design because metric values depend on the actual decisions by optimizers. For example, decisions on which wires take shortest routes and which ones get detoured determine which pairs of wires experience crosstalk and which signal paths become slow \cite{EDA,PDtextbook}. Because of this estimation difficulty, optimization methods with closed-form objectives are fundamentally limited in what they can achieve, and circuit implementation may need to be redone when routing cannot be completed or timing constraints cannot be satisfied \cite{EDA,ABKplacement}.

\noindent
{\bf Key sources.}
To solve mixed-size placement, the Nature paper \cite{Nature} first places macros and then places small components with commercial software. It places macros using a {\em Reinforcement Learning} (RL) action policy that is {\em iteratively improved} ({\em fine-tuned}) at the same time. The RL policy can be {\em pre-trained} on prior circuits or initialized ``{\em from scratch}''.
The iterative process runs for a set time (or until no improvement) and optimizes a fixed (not learned) proxy cost function that blends HPWL, component density, and routing congestion. To evaluate this function, the small components are placed with force-directed placement.
\cite{Nature} claims better results for RL than 3 baselines: ($i$) macro placement by human chip designers, ($ii$) parallel Simulated Annealing, ($iii$) RePlAce software from UCSD, which uses no RL.

Among secondary sources discussed in the context of \cite{Nature} (Table \ref{tab:refs}),
we prefer scholarly papers \cite{ISPD2022,SB,ISPD2023} and draw on open-source repositories and FAQs as needed \cite{CT,MP}. Here all benchmark sets have hundreds of macros per design, compared to only a handful in sets such as ISPD 2015.
We crosscheck claims from three nonoverlapping groups of researchers: those associated with \cite{Nature}, the Stronger Baselines paper \cite{SB} and the UCSD paper \cite{ISPD2023}. Consistent claims from different groups are even more trustworthy when backed by numerous benchmarks. Both Google Team 2 and the UCSD team included highly-cited experts on floorplanning and placement with extensive publication records and several key references cited in \cite{Nature} (such as \cite{MAPLE,Progress,RePlAce} and others), as well as experience developing academic and commercial floorplanning and placement tools beyond Google.

\vspace{-2mm}
\section{Initial doubts}
\label{sec:initial}
\vspace{-1mm}
While the Nature paper \cite{Nature} was sophisticated and impressive, its research plan had notable shortfalls. For one, the proposed reinforcement learning (RL) was presented as being capable of broader combinatorial optimization
(a field that includes puzzle-like tasks such as the Traveling Salesperson Problem, Vertex Cover, Bin Packing).
But instead of illustrating this with key problem formulations and easy-to-configure test examples \cite{RLcomb}, it solved a specialty task (macro placement for chip design) for proprietary Google TPU circuit design blocks, giving results on 5 blocks out of many more available. The RL formulation did not track chip metrics and optimized a simplified {\em proxy} function that included HPWL (see Section \ref{sec:background}), but was not evaluated for pure HPWL optimization on open circuit examples, as is routine in the literature \cite{Capo,Benchmarks,Routable04,FSmixed,FS5,Combinatorial,PDhandbook,CongNam,Cong2013,Progress,RePlAce,ABKplacement}. New ideas in placement are usually evaluated in research contests on industry chip designs released as public benchmarks \cite[Section 6.1]{ABKplacement}, e.g., \cite{CongNam,ICCAD14contest,ICCAD15contestA,ICCAD15contestB,Progress,ICCAD16contest}. but \cite{Nature} neglected these contest benchmarks.

Some aspects of \cite{Nature} looked suspicious, as it ($i$) did not substantiate several claims and withheld key aspects of experiments, ($ii$) claimed improvements in noisy metrics that the proposed technique did not optimize,
($iii$) relied on techniques with known handicaps that undermined performance in similar circumstances, and ($iv$) may have misconfigured and underreported its baselines. We spell these out in Sections \ref{sec:reporting}-\ref{sec:baselines} --- confirming even a fraction of specific concerns  would put the top-line claims and conclusions of \cite{Nature} in serious doubt.

\subsection{Unsubstantiated claims and \\ insufficient reporting}
\label{sec:reporting}
  Several significant omissions can be understood by readers without background in chip design.

\noindent
{\bf U1.}
With ``fast chip design'' in the title \cite{Nature}, the authors only described improvement in design-process time as ``days or weeks'' to ``hours'' without giving per-design time or breaking it down into stages. It was unclear if ``days or weeks'' for the baseline design process included the time for functional design changes, idle time, inferior EDA tools, etc.

\noindent
{\bf U2.}
The claim of RL runtimes {\em per testcase} \cite[Abstract]{Nature} being under six hours (for each of 5 TPU blocks) excluded RL pre-training on 20 blocks (not amortized over many uses, as in some AI applications). Pausing the clock for pre-training (not used by prior methods) was misleading. Also, RL runtimes only cover macro placement, but RePlAce and industry tools place all circuit components.

\noindent
{\bf U3.}
\cite{Nature} focused on placing macros, but did not give the number, sizes or shapes of macros in each TPU chip block used, nor area utilization, etc.

\noindent
{\bf U4.}
\cite{Nature} gave results on only five TPU blocks, with unclear statistical significance, but high-variance metrics produce noisy results (Table \ref{tab:metrics}). Using more examples is common (Table \ref{tab:refs}).%

\noindent
{\bf U5.}
\cite{Nature} was silent on the qualifications and the level of effort of the human chip designer(s) outperformed by RL. Reproducibility aside, those results could be easy to improve (as shown in \cite{ISPD2023} later).

\noindent
{\bf U6.} \cite[Abstract]{Nature} claimed improved ``area'', but chip area and macro area did not change in \cite{Nature}, and standard-cell area did not change during placement (also see the 0.00 correlation in Table \ref{tab:metrics}).

\noindent
{\bf U7.} For iterative optimization algorithms that improve results over time, fair comparisons show {\em per testcase}: ($i$) better quality metrics with equal runtime, ($ii$) better runtime with equal quality or ($iii$) wins for both. \cite{Nature} offered no such evidence. In particular, if ML-based optimization is used with extraordinary compute resources, then so should be optimization by Simulated Annealing in its most competitive form.

\subsection{A flawed optimization proxy}
The chip design methodology in \cite{Nature}
uses physical synthesis to generate circuits for further layout optimization (physical design). The proposed RL technique places macros of those circuits to optimize a simplified {\em proxy cost function}. Then, a commercial EDA tool is invoked to place the remaining components (standard cells). The remaining operations (including power-grid design, clock-tree synthesis and timing closure \cite{PDhandbook,PDtextbook}) are outsourced to an unknown third party~\cite{Nature,PeerReviews}. Results are evaluated with respect to routed wirelength, area, power, and two circuit-timing metrics: TNS and WNS.\footnote{TNS = Total Negative Slack, WNS = Worst Negative Slack. These metrics measure violations of timing constraints (negative {\em slack} represents violations) by adding violations along all critical paths or using the worst violations. These metrics are noisy since chip timing is often determined by a handful of paths, and small changes to macro locations may change timing
a lot.}
Per \cite{Nature}, the proxy cost function did not perform circuit timing analysis~\cite{PDtextbook} needed to evaluate TNS and WNS.\footnote{Proxy values correlate poorly with TNS and WNS
\cite{ISPD2023}.} Therefore, it was misleading to claim in \cite{Nature} that the proposed RL method led to TNS and WNS improvements on five TPU design blocks {\em without} performing variance-based statistical significance tests
(TNS and WNS were optimized at later steps unrelated to RL \cite{Nature}).

\subsection{Use of handicapped techniques}
\vspace{-2mm}
 To experts, the proposed methodology {\em looked} handicapped: using outdated methods made it harder to improve State of the Art (SOTA).

\noindent
{\bf H1.}
 The proposed RL used exorbitant CPU/GPU resources compared to SOTA. Hence, claimed ``fast chip design'' (presumably, due to fewer unsuccessful design attempts) required careful substantiation.

\noindent
{\bf H2.}
Placing macros one by one (a type of {\em constructive}  floorplanning \cite{PDtextbook}) is one of the simplest approaches. Simulated Annealing can swap and shift macros, and make other incremental changes. Analytical methods move many components at once. One-by-one placement looked handicapped even when driven by Deep Reinforcement Learning.

\noindent
{\bf H3.}
\cite{Nature} used circuit-partitioning (clustering) methods
similar to partitioning-based methods from 20+ years ago~\cite{Capo,FS5,Floorplacement,Combinatorial,PDhandbook,PDtextbook}. Those techniques are known to diverge from interconnect optimization objectives~\cite{Combinatorial,PDtextbook}. By placing macros using a clustered netlist {\em without gradual layout refinement}, RL runs into the same problem.

\noindent
{\bf H4.}
\cite{Nature} limited  macro locations to a coarse grid, but SOTA methods~\cite{RePlAce} do not impose such a constraint. Figure \ref{fig:ibm10} illustrates the difference. Even if RL can run without gridding, it might not scale to large enough circuits without coarse gridding.

\noindent
{\bf H5.}
The use of force-directed placement from the 1980s \cite{FD84} in \cite{Nature} left much room for improvement.\footnote{In \cite{DP}, Google Team 1 used a modern method (DREAMPlace \cite{DREAM} derived from RePlAce) instead of force-directed placement but claimed improvement only in proxy costs, not chip metrics.}

\vspace{-1mm}
\subsection{Questionable baselines}
\label{sec:baselines}
\vspace{-2mm}
The Nature paper~\cite{Nature} used several baselines to claim the superiority of proposed techniques. We already mentioned in Section \ref{sec:reporting} that the human baseline was undocumented and not reproducible.

\noindent
{\bf B1.}
Key results of \cite{Nature} report in \cite[Table 1]{Nature} chip metrics for five TPU design blocks. But comparisons to SA do not report those chip metrics.

\noindent
{\bf B2.}
\cite{Nature} mentions that SA was used to postprocess the results of RL, but gives no ablation studies to evaluate the impact of SA on chip metrics.

\noindent
{\bf B3.}
RePlAce \cite{RePlAce} was used as a baseline in \cite{Nature} in a way inconsistent with its intended use. As Section \ref{sec:background} explains, analytical methods do well on circuits with millions of movable components, but RePlAce was not intended for clustered netlists with a reduced number of components --- it should be used directly sans clustering (for details, see \cite{RePlAce,SB,ISPD2023}). Clustering can worsen results due to a mismatch between placement and partitioning objectives~\cite{Combinatorial}, and by unnecessarily creating large clusters that are hard to pack without overlaps.

\noindent
{\bf B4.} \cite{Nature} did not describe how macro locations in SA were initialized, hinting that the naive approach in \cite{Nature} could be improved.
Later, \cite{SB} identified more handicaps in the SA baseline in \cite{Nature},
and \cite{ISPD2023} confirmed them (Section \ref{sec:additional}).

\vspace{-1mm}
\section{Additional evidence}
\label{sec:additional}

 Months after the Nature publication \cite{Nature}, more data became available in \cite{PeerReviews,CT,ISPD2022,SB}, followed by the first wave of controversial media coverage \cite{NYT,Reuters,Wired}.
 Nature editors released the peer review file for \cite{Nature} including authors' rebuttals: in the lengthy back-and-forth with reviewers~\cite{PeerReviews} the authors assured reviewers that macro locations were not modified after placement by RL, confirming coarse-grid placement of macros. Among several contributions, \cite{SB} implemented the request of Nature Reviewer \#3 \cite{PeerReviews} and benchmarked Google's technique on 17 public chip-design examples \cite{ICCAD04}: prior methods decisively outperformed Google RL.  Professors Patrick Madden (SUNY Binghamton) and Jens Lienig (TU Dresden) publicly expressed doubts about the Nature paper~\cite{NYT,Reuters}. As researchers noted gaps in the Google open-source release \cite{CT}, such as the grouping (clustering) flow, Google engineers released more code (but not all), prompting more questions (see below). Another year passed, and \cite{MP,ISPD2023} expanded on the initial suspicions in several ways. They demonstrated that not limiting macro placement allows human designers and commercial EDA tools (separately) to outperform results produced by Google code \cite{CT}. \cite[Table 2]{ISPD2023}
 estimated rank correlation of the proxy cost function optimized by RL to chip metrics used in \cite[Table 1]{Nature}, and
 \cite[Table 3]{ISPD2023} estimated the mean and standard deviation for chip metrics after RL-based optimization. We give a summary in Table \ref{tab:metrics}, where rank correlations are low for all chip metrics, while TNS and WNS are noisy. Hence, the optimization of TNS and WNS in \cite{Nature} relied on a flawed proxy and produced results in \cite[Table 1]{Nature} of dubious statistical significance. We note that $\sigma/|\mu| > 0.5$ for TNS on Ariane-NG45 (also on BlackParrot-NG45 in \cite[Table 3]{ISPD2023}).
 In the second round of critical media coverage, \cite{Nature} was questioned by Profs. William Swartz (UT Dallas), Patrick Madden (SUNY Binghamton), and Moshe Vardi (Rice) \cite{TheRegister,CACM}.

 \begin{table*}
 \vspace{-4mm}
 \begin{center}
 \begin{tabular}{l|c|c|c|c|c}
Chip metrics $\rightarrow$ & area & routed wirelength & power &  WNS & TNS \\
\hline
Rank correlation to RL proxy cost & \bf 0.00 &  \bf 0.28 & \bf 0.05 & \bf 0.20 & \bf 0.05 \\
Mean $\mu$ & 247.1K & 834.8 & 4,978 & -100 & -65\\
Standard deviation $\sigma$ & 1.652K & 4.1 & 272  & 28 & 36.9  \\
$\sigma/|\mu|$ & 0.01 & 0.00 & 0.05 & \bf 0.28 & \bf 0.57 \\
\hline
 \end{tabular}
\vspace{-1mm}
 \caption{
 \label{tab:metrics}
  Evaluating the soundness of the proxy cost used with RL in \cite{Nature} and the noisiness of reported chip metrics after RL-based optimization. We summarize data from \cite[Table 2]{ISPD2023} on the Kendall rank correlation of chip metrics to the RL proxy cost and from \cite[Tables 3 and 4]{ISPD2023} on statistics for chip metrics (only Ariane-NG45 design data are shown, but data for BlackParrot-NG45 shows similar trends). As expected, purely-additive metrics (standard-cell area, routed wirelength and chip power) exhibit low variance, but the TNS and WNS metrics, that measure  timing-constraint violations, have high variance.
 }
 \end{center}
\vspace{-6mm}
 \end{table*}

\subsection{Methods}
 \vspace{-1mm}
 \noindent
{\bf Undisclosed use of $(x,y)$ locations from commercial tools.}
Strong evidence and confirmation by Google engineers are mentioned in the UCSD paper \cite{ISPD2023} that \cite{Nature} withheld a critical detail. When clustering the input netlist, {\sc CT merge} code in \cite{CT} read in a placement to restructure clusters based on locations. To produce $(x,y)$ locations of macros, \cite{Nature} used initial $(x,y)$ locations of all circuit components (including macros!) produced by commercial EDA tools from Synopsys \cite[Issue \#25]{CT}.
The lead authors of \cite{Nature} confirmed
using this step, claiming it was unimportant \cite{AAstatement}. But it improved key metrics by 7-10\% in \cite{ISPD2023}. So, the results in \cite{Nature} needed algorithmic steps absent in \cite{Nature}, e.g., obtaining $(x,y)$ data from commercial software.

\noindent
{\bf More undocumented techniques} were itemized in \cite{ISPD2023}, which mentioned discrepancies between the Nature paper \cite{Nature}, their source code \cite{CT} and the actual code used for chip design at Google. These discrepancies included specific weights of terms in the proxy cost function, a different construction of the adjacency matrix from the circuit, and several ``blackbox'' elements of \cite{CT} available as binaries
with no source code or full description in \cite{Nature}.
\cite{SB} and \cite{MP,ISPD2023} offer missing descriptions.
Moreover, the results in \cite{Nature} did not match the methods in \cite{Nature} because key components were missing. And neither results nor methods were reproducible from descriptions alone.

\noindent
{\bf Data leakage between training and test data?}
Per \cite{Nature}, ``as we expose the policy network to a greater variety of chip designs, it becomes less prone to overfitting.'' But Google Team 1 showed later in \cite[Figure 7]{ISPD2022} that pre-training on ``diverse TPU blocks'' did not improve quality of results. Pre-training on ``previous netlist versions'' improved quality somewhat. Pre-training RL and evaluating it on similar designs \cite{pretraining} might be a serious flaw in methodology of \cite{Nature}.\footnote{Such a methodology could help chip designers iterate on design changes faster, but that was not described in \cite{Nature}.} As Google did not release proprietary TPU designs or per-design statistics, we cannot compare training and test data.

\noindent
{\bf A middling Simulated Annealing baseline.}
The Stronger Baselines paper \cite{SB} from Google Team 2
improved the parallel SA used by Google Team 1 in \cite{Nature}
by adding ``move'' and ``shuffle'' actions to ``swap'', ``shift'' and ``mirror'' actions. This improved SA typically produces better results than RL in a shorter amount of time when optimizing the same objective function. \cite{ISPD2023} reproduced the conclusions of \cite{SB} with an independent implementation of SA and found that SA results had less variance than RL results. Additionally, \cite{SB} suggested a simple and fast macro-initialization heuristic for SA and equalized compute times when comparing RL to SA. Given that SA was widely used in the 1980s and 1990s, not implementing a strong enough SA baseline contributed to wrong conclusions about the superiority of the new RL technique.

\subsection{Results}

\noindent
{\bf Inconsistencies in claimed runtimes.}
\cite{Nature} claims runtimes under six hours, but papers and presentations by Google Team 1~\cite{JD2020,YJLee} reuse the blurred green-blue-white chip image in \cite[Extended Data Figure 5]{Nature} with 12-24 hour runtimes and identical total wirelength (55.42m).

\noindent
{\bf An inconclusive testcase.} Google's RL code \cite{CT} lost to prior methods on most chip-design examples in \cite{SB,ISPD2023} except for Ariane --- the only example released in support of \cite{Nature}. But when macros of Ariane were shuffled \cite{ISPD2023}, chip metrics moved very little. Thus, Ariane was not a challenging testcase.

\subsection{Likely imitations}
\cite{Nature} did not disclose major limitations of its methods but promised success in broader combinatorial optimization. The Ariane design image in \cite[Extended Data Figure 4]{Nature} shows macro blocks of identical sizes: a potential limitation. Yet, \cite{Nature} does not report basic statistics per TPU block: the number of macros and their shapes, design area utilization, and the fraction of area taken by macros. Based on \cite[page 43]{PeerReviews} and the guidance from Google engineers to the authors of \cite{ISPD2023}, it appears TPU blocks had area utilization on the order of 60\%. Commercial chips are often denser, and can use macros of different sizes. Poor performance of Google RL on challenging public benchmarks from \cite{ICCAD04} used in \cite{SB,ISPD2023} (illustrated in Figure \ref{fig:ibm10}) suggests undisclosed limitations. Another possible limitation is poor handling of preplaced (fixed) macros, common in industry layouts, but not discussed in \cite{Nature}. By interfering with preplaced macros, gridding (see H4) can impact usability in practice. Poor performance on public benchmarks may also be due to overfitting to  proprietary TPU designs.

\begin{table*}
\begin{center}
\begin{tabular}{c|c|c|c}
$\downarrow$
Designs / Tools $\rightarrow$  & Google CT/RL & Cadence CMP   &  UCSD SA \\
\Xhline{2\arrayrulewidth}
Ariane-NG45      &  32.31      & 0.05 &  12.50 \\
BlackParrot-NG45 & 50.51       & 0.33 &  12.50 \\
MemPool-NG45     &  81.23      & 1.97 &  12.50 \\
\hline
\end{tabular}
\caption{\label{tab:runtime} Runtimes in hours for three mixed-size placement tools and methodologies on three large chip modern designs reported in the arXiv version of \cite{ISPD2023}. Google CT - Circuit Training code supporting RL in the Nature paper, used without pre-training.
Cadence CMP - Concurrent Macro Placer (commercial EDA tool). SA - parallel Simulated Annealing implemented at UCSD following \cite{SB} given 12.5 hours of runtime in each case. CT and SA are used only to place macros, the remaining components are placed by a commercial EDA tool whose runtime is not included. Cadence CMP places all circuit components.
By quality of results in \cite{ISPD2023} (not shown here), Cadence CMP leads, followed by Simulated Annealing, followed by Google CT. \cite{MP} additionally evaluated Cadence CMP versions by year and concluded that performance and runtime on these examples
did not appreciably change  between the versions since 2019.%
}
\end{center}
\vspace{-4mm}
\end{table*}

\section{Did [1] improve SOTA?}
\label{sec:SOTA}
\vspace{-2mm}

The Nature editorial \cite{NatureEditorial2021} discussing \cite{Nature} speculated that ``this is an important achievement and will be a huge help in speeding up the supply chain.'' But today, after evaluations and reproduction attempts at multiple chip-design and EDA companies, it is safe to conclude that no important achievement occurred because prior chip-design software, particularly from Cadence Design Systems, produced better layouts faster \cite{MP,ISPD2023}.
If this were known to the reviewers of \cite{Nature} or to the public, the claims of improving TPU designs in \cite{Nature} would be nonsensical.
\cite{Nature} claimed that humans produced better results than commercial EDA tools, but gave no substantiation. When license terms complicate publishing comparisons to commercial EDA tools,\footnote{The lawsuit~\cite{FAC,BloombergJuly23} alleges that Google obtained better results with commercial tools before Nature submission.}
one compares to academic software and to other prior methods, with the proviso that small improvements are not compelling.

\vspace{-2mm}
\subsection{Reproduction attempts}
Google Team 2 \cite{SB} and the UCSD Team \cite{MP,ISPD2023} took different approaches to comparing methods from \cite{Nature} to baselines, but cumulatively reported comparisons to commercial EDA tools, to human designers, to prior university software, and to two independent custom implementations of Simulated Annealing.
\begin{itemize}
\vspace{-1mm}
\item Google Team 2 in \cite{SB} followed the descriptions in \cite{Nature} and did not supply initial placement information. The UCSD Team in \cite{MP,ISPD2023} sought to replicate what Google {\em actually} used to produce results (without description in \cite{Nature}).%
\vspace{-1mm}
\item Google Team 2 had access to TPU design blocks and evaluated the impact of pre-training in \cite{SB}. The impact was small at best.\footnote{A consistent conclusion was reported in \cite[Figure 7]{ISPD2022} by Google Team 1 --- training on diverse designs does not improve quality of results, and improvements are seen only when training on earlier versions of the same design. In December 2022 Dr. Jeff Dean, a then Google SVP and the most senior author of \cite{Nature}, also confirmed that RL did well without pre-training \cite{JD}.}%
\vspace{-1mm}
\item The UCSD Team \cite{MP,ISPD2023} lacked access to Google training data and code but followed Google instructions in \cite{CT} for obtaining results similar to those in \cite{Nature} without pre-training. They also reimplemented SA following instructions in \cite{SB} and introduced several new chip-design examples (Table \ref{tab:refs}).
\vspace{-1mm}
\item The UCSD Team \cite{MP,ISPD2023} but not Google Team 2 \cite{SB} performed comparisons using chip metrics and using a commercial EDA tool (Cadence CMP), where Cadence CMP outperfomed Google RL. When running RePlAce in this context, \cite{ISPD2023} used only macro locations produced by RePlAce and placed standard cells with the same commercial software used after Google CT/RL \cite{Nature,CT} (more details below).%
\vspace{-1mm}
\item The UCSD Team \cite{MP,ISPD2023} repeated SA vs. RL comparisons for several configurations (those in \cite{Nature},
those in \cite{CT}, and additional ones suggested by Google engineers). The results were consistent.%
\vspace{-1mm}
\item The UCSD Team \cite{MP,ISPD2023} demonstrated that a chip designer from IBM outperformed Google RL \cite{CT}, whereas Google Team 2 \cite{SB} did not use human baselines.
\vspace{-1mm}
\end{itemize}

For comparisons that can be crosschecked, the two teams report (in \cite{SB} and \cite{MP,ISPD2023}) qualitatively similar conclusions.

As pointed out in \cite{SB}, RePlAce was used in \cite{Nature} in a way inconsistent with its intended use. As a mixed-size placer, RePlAce expects a circuit netlist with macros and standard cells, as a large number of separate components. Instead, the comparisons in \cite{Nature} suppressed the advantage of RePlAce by clustering standard cells into a few large chunks.\footnote{Per Section \ref{sec:background}, analytical placers like RePlAce \cite{RePlAce} beat other methods on circuits with millions of components. With $<$100K components, earlier methods can be competitive.}  With proper use of RePlAce, \cite{SB} and, independently, \cite{ISPD2023} obtain strong results with RePlAce on well-known public ICCAD 2004 benchmarks.\footnote{Comparing HPWL results to those in \cite{FSmixed,Combinatorial}, Google CT/RL~\cite{CT} underperforms Feng Shui \cite{FS5} circa 2005.} In comparisons on recent designs \cite{MP,ISPD2023}
RePlAce also attains better chip metrics, but still loses to Google CT/RL because its placements are harder to route (the losses are much smaller than those reported in \cite{Nature}). Notably, \cite{SB,MP,ISPD2023} used RePlAce in a fast mode and not high-quality mode. Congestion-driven mode in RePlAce \cite{RePlAce} was not used. In contrast,
\cite{Nature} used routability-improvement techniques, such as cell bloating, without disclosing them (according to \cite{ISPD2023}). Such techniques can be combined with RePlAce to ensure fair comparisons. Other techniques~\cite{Routable04,PDhandbook,Cong2013,Progress} can be used too.

As explained in \cite{SB}, the implementation of Simulated Annealing used in \cite{Nature} was handicapped. Removing the handicaps (in the same source code base) improved results. When properly implemented, SA produces better solutions than Google CT/RL \cite{CT} using less runtime, when both are given the same proxy cost function. This is shown consistently in \cite{SB,ISPD2023} on 17 widely used ICCAD 2004 benchmarks \cite{ICCAD04} and in \cite{ISPD2023} on several modern design benchmarks.
Compared to Google CT/RL \cite{CT}, SA consistently improves wirelength and power metrics. For circuit timing metrics TNS and WNS, SA produces less noisy results that are otherwise comparable to RL's results \cite{ISPD2023}. Recall that the proxy function optimized by
SA and RL does not include timing metrics \cite{Nature},
making any claims of improvement in these metrics due to SA or RL dubious.

Improving upon SOTA requires improving upon all prior baselines.
Google CT/RL failed to improve by quality upon ($i$) human baselines, ($ii$) commercial EDA tools, and ($iii$) SA. It did not improve SOTA by runtime either (Table \ref{tab:runtime}), and \cite{Nature} did not disclose per-design data or design-process time.
RePlAce and SA gave stronger baselines than described in \cite{Nature},
when configured/implemented well.

\subsection{Open contest at MLCAD 2023}

The EDA industry and research community regularly organize open research contests to keep track of SOTA on industry-produced chip designs, to evaluate published methods and codes, and to perform fair comparisons. A recent contest is relevant to our meta-analysis.
In 2023, the IEEE/ACM Workshop on Machine Learning in Computer-Aided Design (MLCAD) held an open research contest for macro placement for FPGAs and ASICs ({\blue{\url{https://mlcad-workshop.org/1st-mlcad-contest/}}).
The contest followed the playbook established at other CAD conferences such as DAC, ICCAD, and ISPD, and particularly successful in evaluating algorithmic techniques for physical design since the ISPD 2005 placement contest~\cite{CongNam,ICCAD14contest,ICCAD15contestA,ICCAD15contestB,Progress,ICCAD16contest}:
\begin{itemize}
\vspace{-1mm}
\item a challenge problem is announced along with chip design examples and metrics;
\vspace{-2mm}
\item several months after sign-up, final submissions are collected and evaluated {\em by the organizers} on hidden design examples (benchmarks); the same compute resources are available to all contestants (fairness);%
\vspace{-2mm}
\item after evaluation, all benchmarks and winning solutions are released in public;
\vspace{-2mm}
\item  to facilitate industry participants, source code release is typically not required to win, but rather encouraged by additional prize money.
\vspace{-1mm}
\end{itemize}

The MLCAD contest focused on the macro placement task ``inspired by recent deep reinforcement learning (RL) approaches (e.g. [1]),'' aiming ``to improve upon the current state-of-the-art macro placement tools.'' Compared to \cite{Nature}, macro placements had to satisfy additional constraints. The objective function minimized during the contest did not include circuit-timing evaluation, just as the RL approach in \cite{Nature} did not. Numerically, the objective multiplied penalty terms for the runtime of macro placement, the runtime of downstream place-and-route tasks and the resulting routing congestion evaluated on a grid.\footnote{
The inclusion of runtime in the objective function (common to contests in physical design) was likely the most significant difference from evaluation in \cite{Nature},
where large amounts of parallel computational resources were used in comparison to prior methods that used smaller computational resources. In industry practice, design-process time is important, and the authors of \cite{Nature} advertised design-process speed in the title.} Completing macro placement in under 10 minutes led to no penalty for runtime, and this was the case for many teams on many benchmarks. At the same time, poor macro placements tend to lead to high routing congestion and long place-and-route runtimes. Results from multiple design examples were combined using the geometric mean.
The organizers first provided a public benchmark dataset of 140 designs with numerous macros and varying levels of difficulty, and then used a separate set of 198 ``hidden'' designs for evaluation (cf. results on only five design blocks in \cite{Nature}). All designs were released after the contest results were announced.

Despite technical differences from infrastructure in \cite{Nature}, the contest offered a great opportunity for Google and the authors of \cite{Nature} to showcase the versatility and quality of their RL technique.\footnote{RL was pitched in \cite{Nature} as a general technique for combinatorial optimization, so must handle various macro-placement tasks.} The contest attracted 19 participants, of them 8 finalists --- academic teams from Taiwan, Hong Kong, China, Canada and the US (several students per team). Google did not take part.\footnote{
Google did not skip a similar research contest at the Intl. Workshop on Logic Synthesis (IWLS 2023). That contest (\blue{ \url{https://github.com/alanminko/iwls2023-ls-contest}}) focused on chip logic design rather than physical design (two nearby fields). Google won the first place using long-running parallel Simulated Annealing (but during the IWLS 2024 Contest another team outperformed Google results without using ML or DNNs, per \blue{\url{https://github.com/alanminko/iwls2023-ls-contest}}). The winning team overlapped with Google Team 1 working on macro placement: Sergio Guadarrama (quoted in \cite{WSJ-IWLS}), the senior staff software engineer co-authored the ISPD 2022 paper \cite{ISPD2022} with the lead authors of \cite{Nature} and then tweeted on 5/3/22 that he and his team at Google ``independently replicated'' the work in \cite{Nature}.}

Contest results were announced on September 13, 2023. According to the participants' own descriptions, top six teams used traditional analytical optimization methods sans ML --- based on either DREAMPlace \cite{DREAM} (derived from RePlAce) or SimPL \cite{SimPL}. The absence of RL solutions was noteworthy, given that Google ``open-sourced'' the methods of \cite{Nature} in \cite{CT}.
Evaluation scores on hidden benchmarks generally mirrored the trends seen on public benchmarks, which is not surprising in the absence of ML methods (that sometimes overfit to training data). Overall contest results were consistent with the conclusions in \cite{SB,ISPD2023} that traditional optimization methods in circuit placement produced better macro placement results than RL from \cite{Nature} did, and finished faster. The results in \cite{ISPD2023} complete the overall picture by evaluating chip metrics used in \cite{Nature}: power, circuit delay, and the total area of standard cells. Industry chips are typically evaluated using the PPA triad (Power, delay Performance, and chip Area), but macro placement methodologies considered do not alter chip area.

\section{Rebuttals to critiques of [1]}
\label{sec:responses}

 Despite critical media coverage \cite{NYT,Reuters,TheRegister,IEEESpectrum} and technical questions raised, the authors of \cite{Nature,CT} failed to remove remaining obstacles to reproducibility~\cite{CACM} of the methods and results in \cite{Nature}. The UCSD team's engineering effort overcame those obstacles, and they followed up on the work of Google Team 2 \cite{SB} that criticized \cite{Nature}, then analyzed many issues listed in Sections \ref{sec:initial} and \ref{sec:additional}.
Google Team 2 had access to Google TPU designs and the source code used in \cite{Nature} before the CT GitHub repo \cite{CT} appeared. The UCSD authors of \cite{MP,ISPD2023} had access to Circuit Training (CT) \cite{CT} and benefited from a lengthy involvement of Google Team 1 engineers, but had no access to SA code used in \cite{Nature,SB} or other key pieces of code missing from \cite{CT}. Yet, the results in \cite{SB} and \cite{MP,ISPD2023}
corroborate each other, and their qualitative conclusions are consistent.
UCSD results for Ariane-NG45 closely match those by Google Team 1 engineers, and
\cite[Figure 4]{ISPD2023} shows that CT training curves of Ariane-NG45 generated at UCSD match those produced by Google Team 1 engineers.  Google Team 1 engineers carefully reviewed the paper \cite{ISPD2023} and the work in Fall 2022 and Winter 2023, raising no objections \cite[FAQ]{MP}.

 The two lead authors of \cite{Nature} left Google in August 2022, but in March 2023 objected to results of \cite{ISPD2023} without remedying the deficiencies of \cite{Nature} (Section \ref{sec:initial}). Those objections were addressed promptly in \cite[FAQ]{MP}, e.g., in \#6, \#11, \#13, \#15. One issue was the lack of {\em pre-training} in experiments in \cite{ISPD2023}.

\noindent
{\bf Pre-training.} \cite{ISPD2023} performed training using code and instructions in Google's Circuit Training (CT) repo \cite{CT}, which states (June 2023): ``The results below are reported for training from scratch, since the pre-trained model cannot be shared at this time.''
\begin{itemize}
\vspace{-2mm}
\item
Per MP FAQ in \cite{MP}, \cite{ISPD2023}  did not use pre-training because, per Google's CT FAQ \cite{CT}, pre-training was not needed to reproduce results of \cite{Nature}. Also, Google did not release pre-training data.%
\vspace{-2mm}
\item Google Team 2 \cite{SB} evaluated pre-training using Google-internal code and saw no impact on comparisons to SA or RePlAce.
\vspace{-2mm}
\item  Google Team 1 showed \cite[Figure 7]{ISPD2022} that pre-training on ``diverse TPU blocks'' did not improve results, only runtime. Pre-training on ``previous netlist versions'' gave small improvement. No such previous versions were discussed, disclosed or released in \cite{Nature,CT}.
\vspace{-2mm}
\item Dr. Jeff Dean's presentation \cite{JD} gave strong RL results ``from scratch'' (w/o pre-training).
\vspace{-2mm}
\end{itemize}
In other words, the lead authors of \cite{Nature} want others to use pre-training while they did not describe it in detail sufficient for reproduction, did not release code or data for it, and have themselves shown that it does not improve results in the context of \cite{Nature}. Pre-training can also be abused \cite{pretraining}.

\noindent
{\bf Old benchmarks}. Another objection \cite{IEEESpectrum} is that public benchmarks \cite{ICCAD04} used in \cite{SB,ISPD2023} allegedly use outdated infrastructure. But those circuits have been evaluated with the HPWL objective, which scales accurately under geometric 2D scaling of chip designs and remains appropriate for all technology nodes (Section \ref{sec:background}). Per \cite{PeerReviews}, ICCAD benchmarks were requested \cite{PeerReviews} by Peer Reviewer \#3 of \cite{Nature}. When \cite{SB,ISPD2023} implemented this ask, Google RL ran into trouble before routing became relevant: RL lost by 20\% or so in HPWL optimization (HPWL is the simplest yet most important term of the proxy cost optimized by CT/RL \cite{Nature,CT}).

\noindent
{\bf Not training until convergence} in experiments in \cite{ISPD2023}. This concern was promptly addressed in FAQ \#15 in \cite{MP}: `` `training until convergence' is not described in any of the guidelines provided by the CT GitHub repo for reproducing the results in the Nature paper.'' The authors of \cite{ISPD2023} followed Google's guidelines in the CT repo \cite{CT}. Later, their additional experiments indicated that ``training until convergence worsens some key chip metrics while improving others, highlighting the poor correlation between proxy cost and chip metrics. Overall, training until convergence does not qualitatively change comparisons to results of Simulated Annealing and human macro placements reported in the ISPD 2023 paper.'' RL-vs.-SA experiments in \cite{SB} predated \cite{CT}, so trained until convergence per the six-hour protocol from \cite{Nature}.

\noindent
{\bf Computational resources} used in \cite{Nature} were very large, costly, and difficult to replicate. Since both RL and SA algorithms produce valid solutions early and then gradually improve the proxy function, the best-effort comparisons in \cite{ISPD2023} used smaller computational resources than in \cite{Nature}, with parity between RL and SA. The result: SA beat RL. \cite{SB} compared RL to SA using the same amount of computational resources as in \cite{Nature}. Their results were consistent with \cite{SB}. If given greater resources, SA and RL are unlikely to further improve chip metrics due to poor correlation to the proxy function from \cite{Nature}.

The lead authors of \cite{Nature} mention in \cite{AAstatement} that \cite{Nature} is heavily cited, but cite no positive reproductions outside Google that cleared all known obstacles. \cite{SB,ISPD2023} do not discuss other ways (than in \cite{Nature,CT}) to use RL in IC design, so we avoid general conclusions.

\section{Can the work in [1] be used?}
\label{sec:use}

The Nature paper \cite{Nature} claimed applications to recent Google TPU chips, providing credence to the notion that those methods improved State of the Art. But aside from vague general claims,
no chip-metric improvements were reported for specific production chips.\footnote{\cite[Table 1]{Nature} shows results for TPU designs of an earlier generation (that is, on chips that were already manufactured at the time).
Assuming substantial use in production, more recent TPU design blocks must have used \cite{Nature,CT} for tape-out.}
Section \ref{sec:SOTA} shows that the methods of \cite{Nature,CT} lag behind Simulated Annealing from the 1980s \cite{SA83,TW85,ParAnneal87,SAbook88}. Moreover,
the Google-internal implementation of SA from \cite{SB} (or the public one from \cite{ISPD2023,MP}) could serve as a drop-in replacement of RL in \cite{Nature,CT}. Without inside knowledge, we speculate how to reconcile the claimed use in TPUs with Google CT/RL lagging behind SOTA (per \cite{SB,ISPD2023}).

\begin{itemize}
\vspace{-1mm}
\item Given the high variance of chip-timing metrics TNS and WNS in RL results (due to low correlation with the proxy metric), trying many independent randomized attempts with variant proxy cost functions and hyperparameter settings may improve best-seen results \cite{FAC}, with much greater runtimes. But SA can also be used this way.%
\vspace{-1mm}
\item Using in-house methods, even if inferior, is a common methodology in industry practice called {\em dogfooding} (``eat your own dogfood'').
Blocks that are not critical (do not affect chip speed) are good candidates for dogfooding. This can explain selective ``production use'' and reporting.%
\vspace{-1mm}
\item Per \cite{Nature}, the results of RL were postprocessed by SA but the CT FAQ \cite{CT} disclaimed this postprocessing --- postprocessing was used in the TPU design flow but not when comparing RL to SA. But since full-fledged SA consistently beats RL \cite{SB,ISPD2023}, SA could substitute for RL (initial locations can be used with {\em adaptive temperature scheduling} in SA).
\vspace{-1mm}
\item Google Team 1's follow-up \cite{ISPD2022} shows (in Figure 7) that pre-training improves results only when pre-training on essentially the same design. Google could be using RL when performing multiple revisions to IC designs --- a valid context, but not covered in \cite{Nature}. Besides, commercial EDA tools are orders of magnitude faster (running from scratch) than RL  (Table \ref{tab:runtime}), so pre-training RL does not close the gap with \cite{Nature}.%
\vspace{-1mm}
\item Per \cite{PeerReviews,ISPD2023}, TPU blocks exhibit much lower area utilization during placement (roughly 60\%) than is common in modern chips.
Configuring EDA tools for this context can be challenging. Court materials \cite{FAC} indicate that misleading comparisons due to misconfigured EDA tools were flagged at Google but not corrected.
\end{itemize}

\noindent
{\bf Can Google CT/RL code \cite{CT} be improved?}
RL and SA are orders of magnitude slower than SOTA (Table \ref{tab:runtime}), but pre-training (missing in CT) speeds up RL \cite[Figure 7]{ISPD2022} by only several times.

The CT repository \cite{CT} now contains attempted improvements (such as upgrading~\cite{DP}  force-directed placement~\cite{FD84} to DREAMPlace~\cite{DREAM}), but we have not seen serious improvements to chip metrics. Four major barriers to improving \cite{Nature,CT} remain:
\begin{enumerate}
\vspace{-1mm}
\item The proxy cost function optimized by RL does not reflect circuit timing \cite{ISPD2023}, so improving RL may not help to improve TNS and WNS.
\vspace{-1mm}
\item \cite{SB,ISPD2023} show that SA outperforms RL when optimizing a given proxy function. Hence, RL may lose even with a better proxy.
\vspace{-1mm}
\item RL's placement of macros on a coarse grid limits their locations (Figure \ref{fig:ibm10}). When a human designer ignored the coarse grid, they found better macro locations \cite{ISPD2023}. Commercial EDA tools also avoid this limitation and outperform Google CT/RL.
\vspace{-1mm}
\item Clustering as a preprocessing step creates mismatches between placement and netlist partitioning objectives~\cite{Combinatorial,PDtextbook}.
\vspace{-1mm}
\end{enumerate}

\section{Conclusions}
\label{sec:conclusions}

\vspace{-2mm}
Electronic Design Automation has been relying on common AI methods for dozens of years, including A*-search for wire routing, SAT-solving and inductive theorem-provers for verification, etc. Attempts to use ML for combinatorial optimization are more recent. This meta-analysis discusses the reproduction and evaluation of results in the Nature paper \cite{Nature} on ML for macro placement, as well as the validity of methods, results and claims in that paper.
In \cite{Nature}, we find a smorgasboard of {\em questionable practices in ML} \cite{Questionable2024} including irreproducible research practices, multiple variants of cherrypicking, misreporting, and likely data contamination (leakage).
Based on crosschecked newer data, we draw conclusions with ample redundancy (resistant to isolated mistakes): the integrity of \cite{Nature} is substantially undermined owing to errors in the conduct, analysis and reporting of its study. Omissions, inconsistencies, mistakes, and misrepresentations impacted methods, data, results and interpretation in \cite{Nature}.
Table \ref{tab:fraud} makes it clear that serious issues were raised at Google about the work many months before the Nature publication \cite{Nature}.

\begin{table*}
\vspace{-3mm}
\begin{center}
\begin{tabular}{cc}
Timeline  &  Excerpts from the August 4, 2023 ruling by Judge Frederick S. Chung  \\
\hline
\hline
\parbox{3cm}{
\strut
\href{https://arxiv.org/abs/2004.10746}{\blue{arXiv:2004.10746}} was published on April 22, 2020 by the authors of \cite{Nature} and flagged by Chatterjee in October 2020.
\strut
}
 &
 \parbox{11cm}{
 \strut
 Chatterjee claims that he believed the arXiv paper was fraudulent on three different levels and attempted to report and correct these issues by presenting his team's findings. [...]
 The FAC further alleges that in October 2020, Chatterjee expressly raised concerns that Google ``could be charged with fraud if it continued to represent'' to third parties or partners for commercial agreements that Google's methods were comparatively better than competitor's.
 \strut
 }
 \\
 \hline
 \parbox{3cm}{
 Chatterjee was rebuffed by Employee Relations in
April 2021.} &
 \parbox{11cm}{
 \strut
 Google's ``Employee Relations'' department contacted Chatterjee, and then on April 15, 2021, it disciplined Chatterjee with a written warning, noting Chatterjee's ``unprofessional tone and manner,'' which included ``making uncredible claims of fraud and academic misconduct." [...]\\
 Communications from the Employee Relations department further corroborate Chatterjee's participation in allegedly protected activity.
  \strut
 } \\
 \hline
\parbox{3cm}{
Nature paper \cite{Nature}
was published in
June 2021.
}&
\parbox{11cm}{
 \strut
The FAC subsequently alleges that the arXiv paper --- defined subsequently as the Nature paper --- was published without the contradicting data or disclaimers in the Nature journal on June 9, 2021. [...] There is no basis for striking this allegation.
 \strut
} \\
 \hline
 \parbox{3cm}{
 Chatterjee urged corrections again in
 {February 2022}.} &
 \parbox{11cm}{
  \strut
  On February 18, 2022 Chatterjee again urged correction of the scientific record, asserting it was ``not only the ethical thing to do, but also the legal thing to do."
  \strut
 }
\\
\hline
\parbox{3cm}{
 Chatterjee was fired in
 {\small March 2022}.}
  &
 \parbox{11cm}{
  \strut
  On March 23, 2022, Chatterjee was fired from Google, because he allegedly threatened to disclose his suspicions of fraud to the CEO and the Board.
  \strut
 }
\\
\hline
\end{tabular}
\vspace{-1mm}
\parbox{13cm}{
\caption{\label{tab:fraud}
Timeline of alleged fraud and scientific misconduct per \cite{RulingAug23}.
FAC refers to \cite{FAC}. ``Protected activity'' refers to whistleblower protections under California law. }
}
\end{center}
\vspace{-4mm}
\end{table*}

\subsection{Conclusions about [1]}
\vspace{-1mm}

We crosscheck the results reported in \cite{SB,MP,ISPD2023} and also account for \cite{PeerReviews,CT,ISPD2022,DP}, then summarize conclusions drawn from these works. This confirms many of the initial doubts about \cite{Nature} and identifies additional deficiencies. As a result, it is clear that \cite{Nature} is misleading in several ways, such that the readers can have no confidence in the top-line claims of \cite{Nature}. nor its conclusions.  \cite{Nature} did not improve SOTA while the methods and results of the original paper were not reproducible from the descriptions provided, contrary to stated editorial policies at Nature (see Section \ref{sec:policy}).
The reliance on proprietary TPU designs for evaluation, along with insufficient reporting of experiments, continues to obstruct reproducibility of the methods and the results. Google Team 2 \cite{SB} had access to Google internal code whereas the UCSD Team \cite{ISPD2023} reverse-engineered and/or reimplemented missing components. Google Team 2 and the UCSD team drew consistent conclusions from similar experiments, and each team made additional observations.

\begin{enumerate}
\item
\cite{Nature} reported improvements in several chip-timing metrics (TNS and WNS) that
were not explicitly tracked or optimized by the proposed RL method, and those metrics did not correlate with the proxy objective
used in optimization~\cite{ISPD2023}. Those timing metrics were optimized in postprocessing by commercial EDA tools.
\vspace{-0.2mm}
\item
 Design-process time improvements over human chip designers --- a key claim of \cite{Nature} --- were not reported {\em per testcase} or detailed, and the humans involved were not documented. Later, it was clarified in the CT FAQ \cite{CT} that those human experts somehow used commercial autoplacement tools. However, \cite{ISPD2023} has shown how Google CT/RL was outperformed, in separate comparisons,  by different human designers and by fully automated commercial EDA tools.
 \vspace{-0.2mm}
\item
As first suggested in \cite{SB} and confirmed in \cite{ISPD2023}, each algorithmic baseline described in \cite{Nature} was lacking in some ways and not difficult to improve. As a result, prior methods outperform the methods of \cite{Nature,CT} in quality and runtime.
\vspace{-0.2mm}
\item
The claim of six-hour runtimes for RL macro placement \cite{Nature} is in doubt because the authors of \cite{Nature} reported at conferences \cite{YJLee,JD} 12- and 24-hr runtimes with the same chip images \cite[Extended Data Figure 5]{Nature} and identical total wirelength.  
Moreover, the Nature authors may have stopped the clock during pre-training, which took much longer than six hours and was not amortized over a large number of designs. Either way, commercial tools run orders of magnitude faster (Table \ref{tab:runtime}).
\item
\vspace{-0.2mm}
\cite{Nature} withholds important details required to produce reported results. One of these details is the use of $(x,y)$ locations produced by commercial software. Using these initial locations with Google's RL technique markedly improves the $(x,y)$ locations produced by it \cite{ISPD2023}.
\item Improving the methods of \cite{Nature} to make them competitive would be challenging due to the four barriers itemized in Section \ref{sec:use}.
\end{enumerate}

\subsection{Conclusions for chip-design tech}

Lacking conclusions about specific chip designs, we focus on chip design technologies.

\begin{enumerate}
\item
Machine-learning from entire chip designs is hard: learning from diverse designs might only improve runtime and not quality in typical cases \cite{ISPD2022}. Learning from earlier versions of the same design can be useful in some cases, but should be compared to warm-starting Simulated Annealing with an initial placement and adaptive temperature schedule.%
\vspace{-1mm}
\item
Gridding and clustering methods (popular 20 years ago, but outperformed by ``flat" methods) do not offer new capabilities at this point.%
\vspace{-1mm}
\item
Using exorbitant CPU/GPU resources in \cite{Nature} did not help outperform SOTA. It only complicated experimentation and reproducibility.%
\vspace{-1mm}
\item
The work in \cite{Nature,CT} made a keen observation that physical synthesis tools produce $(x,y)$ locations usable as initial solutions for mixed-size placement. Sadly, this observation was not disclosed in the text of \cite{Nature} but only used to improve results. As it is not specific to RL, it does not support the superiority of RL \cite{ISPD2023}. On the other hand, initial placements were recently studied in \cite{PlaceInit23} and can be reflected in future placement benchmarking efforts.
\vspace{-1mm}
\item
The modern open-source infrastructure \cite{MP} for evaluating macro placers developed for \cite{ISPD2023} can be used to check new ideas and software.\footnote{The design examples in \cite{MP} roughly match \cite{Nature} in area utilization. Increasing area utilization would create harder benchmarks, keeping in mind that higher area utilization decreases fabrication cost for mass-produced ICs.}
\cite{MP,ISPD2023} included in its evaluation a new ML-based macro placer AutoDMP \cite{AutoDMP} from Nvidia that produced promising results without using reinforcement learning.
At the same time, older circuit benchmarks (such as \cite{ICCAD04}) remain relevant, difficult and practically useful. They circumvent proprietary chip infrastructure
and enable, with minimal effort, quick directional comparisons valid for any technology node.
\item A 2024 effort from China \cite{MPEval2024}
compared seven techniques for mixed-size placement using their new independent evaluation framework with 20 circuits (7 with macros).
End-to-end results for chip metrics show that post-\cite{Nature} ML-based techniques lag behind RePlAce \cite{RePlAce} (embedded in OpenROAD) and other optimization-based techniques: DREAMPlace (a GPU-based variant of the RePlAce algorithm) \cite{DREAM} and AutoDMP (a Bayesian Optimization wrapper around DREAMPlace) \cite{AutoDMP}. Despite the obvous need to replicate the methods from \cite{Nature}, the authors of \cite{MPEval2024} were unable to provide such results.%
\end{enumerate}
When a particular design technology underperforms, this does not
necessarily reflect on the actual IC designs where it was attempted.

\subsection{Policy implications}
\label{sec:policy}

Theoretical arguments and empirical evidence suggest that numerous published papers across various fields cannot be replicated and are likely false~\cite{False,SaganWiki,Repro2015,Errors23,Guardian2023,CellMisconduct2024}. The developments with \cite{Nature} add to the so-called {\em reproducibility crisis} that undermines trust in published research results \cite{Battle23,Guardian2023}. In response to this crisis, some observers say ``reproduce or it didn’t happen'' \cite{Didnthappen2023}.

Retraction Watch now tracks 5000 retractions per year, including prominent cases of research misconduct~\cite{FTfraud,Guardian2023}.
Per \cite{Battle23}, ``research misconduct is a serious problem and (probably) getting worse", which makes it even more important to separate honest mistakes from deliberate exaggerations and misconduct
\cite{Exaggerate23,NYTstanford,FTfraud,Science2023,CellMisconduct2024}.
To this end, see Table \ref{tab:fraud}.
Institutional response is needed, and opportunities for future reforms are discussed in \cite{Errors23,Science2023,NatureChina2024,NatureNegative2024}. Particularly important are faster, more numerous retractions \cite{OranskyNature2022} and clarity in Nature retraction notices \cite{NatureRetractions2024}.
Here we make a modest contribution to this far-reaching discussion by making specific suggestions.

\noindent
{\bf Google should follow Google AI principles}
\url{https://ai.google/responsibility/principles}), in particular, Section ``6. Uphold high standards of scientific excellence" that says:
\begin{displayquote}
``Technological innovation is rooted in the scientific method and a commitment to open inquiry, intellectual rigor, integrity, and collaboration[...] We aspire to high standards of scientific excellence...''
\end{displayquote}

The April 7, 2022 tweet by ex-Head of Google Brain~\cite{ZoubinTweet}
appears to contradict the facts: the work in the Nature paper \cite{Nature} was never fully open-sourced and was not independently reproducible because several key parts were not described in the paper or released in code. This was
stated in \cite{SB} prior to the tweet, obvious from \cite{CT} (and publicly mentioned to the lead authors of \cite{Nature} in March 2022), was later documented in detail in \cite{MP,ISPD2023} and explained in plain English in \cite{CACM}.
The still-underspecified use of \cite{Nature} on Google TPU designs (only on selected blocks? trained and tested on similar blocks?) does not counter strong evidence in \cite{SB,MP,ISPD2023} that \cite{Nature} failed to improve SOTA. Many chips are designed every year without improving SOTA, but prior SOTA improvements did not merit Nature publications.

It is unclear why Google did not allow publishing \cite{SB} (coauthored by the author of this meta-analysis), even after its results and conclusions were corroborated by the published paper \cite{ISPD2023} written at UCSD with lengthy involvement from Google. Hesitation to act on \cite{Nature} is understandable, even after \cite{SB} and \cite{MP,ISPD2023} found major flaws in \cite{Nature}, but ``a commitment to open inquiry, intellectual rigor, integrity, and collaboration'' must protect legitimate research in \cite{SB}.
\ \\

\noindent
{\bf Nature Portfolio editorial policies should be followed  broadly and rigorously.}
Quoting from
\url{https://www.nature.com/nature-portfolio/editorial-policies/reporting-standards}:
\begin{displayquote}
\small
``An inherent principle of publication is that others should be able to replicate and build upon the authors' published claims. A condition of publication in a Nature Portfolio journal is that authors are required to make materials, data, code, and associated protocols promptly available to readers without undue qualifications[...] After publication, readers who encounter refusal by the authors to comply with these policies should contact the chief editor of the journal.''
\end{displayquote}

Specifically for \cite{Nature}, the Nature editorial \cite{NatureEditorial2021} insisted that ``the technical expertise must be shared widely.'' But when manuscript authors neglect requests for public benchmarking and obstruct reproducibility, their technical claims should be viewed with suspicion \cite{CACM} (especially if they later disagree with comparisons to their work \cite{AAstatement}). Per the peer review file \cite{PeerReviews}, the acceptance of the Nature paper was conditional on the release of code and data in the second revision, but this did not happen when the paper was published or later, per \cite{ISPD2023}. The Nature paper \cite{Nature} was amended by the authors to claim that the code had been made available (see the ``Data and Code Availability'' disclaimer). But serious omissions remain in the released code. This is particularly concerning because ($i$) \cite{Nature} omitted key comparisons and details, and ($ii$) fraud was alleged under oath in a California court by a Google whistleblower tasked with evaluating the project \cite{FAC}. This makes reproducibility more critical.

\subsection{Nature editors investigate}

In May 2022, \cite{Reuters} quoted a statement by Nature about \cite{Nature}: ``Issues relating to the paper have been brought to our attention and we are looking into them carefully." In March 2023 \cite{TheRegister} reported that
\begin{displayquote}
\small
``Some academics have since urged Nature to review Google's paper in light of UCSD's study. In emails to the journal viewed by The Register, researchers highlighted concerns raised by Prof. Kahng and his colleagues, and questioned whether Google's paper was misleading.''
\end{displayquote}
Further, ``Nature told The Register it is looking into Google's paper... This process involves consultation with the authors and, where appropriate, seeking advice from peer reviewers and other external experts.'' Soon after, \cite{CACM} made a plain-language case that \cite{Nature} lacked reproducibility.
On September 20, 2023, Nature added a note to \cite{Nature} online \cite{RW2023}:
\begin{displayquote}
\hspace{-1mm}
\small
Editor’s Note: Readers are alerted that the performance claims in this article have been called into question. The Editors are investigating these concerns, and, if appropriate, editorial action will be taken once this investigation is complete.
\hspace{-1mm}
\end{displayquote}

A year later (late September 2024), the Editor's note was removed from the Nature article, but an authors' addendum appeared. The addendum largely repeats the arguments from an earlier statement \cite{AAstatement} which we discussed in Section \ref{sec:responses}, so there was little for us to modify in the present article: none of the major concerns about the Nature paper have been addressed. In particular, "results" on one additional proprietary TPU block with undisclosed statistics do not support any serious conclusions. This only aggravates concerns about {\em cherry picking} and {\em misreporting}. The release of a pretrained model without information about training data aggravates concerns about {\em data contamination} - any public benchmark could have been used in training. We do not comment on the Google blog post,\footnote{\blue{\url{https://deepmind.google/discover/blog/how-alphachip-transformed-computer-chip-design/}}} except that it repeats the demonstrably false claim of a full source code release that allows one to reproduce the results in the Nature paper. Among other pieces, source code for simulated annealing is still missing, and additionally the Nature results cannot be reproduced without proprietary training data and test data.

\eat{
A year later, as of this writing, the investigation has not produced results. And we do not have high hopes, given the Nature News team's detailed account of an unrelated scandal involving two retracted Nature papers on high-temperature superconductivity by Ranga Dias  \cite{NatureScandal2024,NatureScandalDetails2024}.
\begin{displayquote}
\hspace{-1mm}
\small
The retraction does not state what Hamlin and Ramshaw found in the post-publication review process instigated by Nature: that the raw data were probably fabricated.
\hspace{-1mm}
\end{displayquote}
Per Nature editor Karl Ziemelis, ``Allegations of possible misconduct are outside the remit of peer review and more appropriately investigated by the host institution.'' That works poorly: {\em three} Univ. of Rochester investigations ``did not find evidence of misconduct,'' and only the fourth found ``data reliability concerns.''
\begin{displayquote}
\hspace{-1mm}
\small
Publicly, Dias continued to insist that CSH was legitimate and that the retraction was simply down to an obscure technical disagreement.
\hspace{-1mm}
\end{displayquote}
Company-internal investigations have even fewer incentives to disclose misconduct to the public, whereas litigation is usually resolved with private settlements and non-disclosure agreements to avoid public admission of wrongdoing.
}

\begin{center}
\hspace{-2mm}
\parbox{6cm}{\hrulefill}
\hspace{-2mm}
\end{center}

We believe it is in everyone's interest to reach clear and unequivocal conclusions about published scientific claims, free of misrepresentations. Authors, Nature editors and reviewers, and the research community, share the burden of responsibility. Seeking the truth is a shared obligation \cite{NYTstanford,Science2023}. 

\noindent
{\bf Acknowledgments.} This meta-analysis would be impossible without the hard work and dedication to science of the authors of \cite{SB} and \cite{MP,ISPD2023}.

\newcommand{\ISPD}[0]{{\em Proceedings of Intl. Symp. Physical Design}\xspace}
\newcommand{\ICCAD}[0]{{\em Proceedings of Intl. Conf. Computer-Aided Design}\xspace}
\newcommand{\DAC}[0]{{\em Proceedings of Design Automation Conf.}\xspace}
\newcommand{\TCAD}[0]{{\em IEEE Transactions on Computer-Aided Design of Integrated Circuits}\xspace}

\end{document}